\documentclass[conference]{IEEEtran}

\usepackage{cite}
\usepackage{amsmath,amssymb,amsfonts}
\usepackage{graphicx}
\usepackage{xcolor}
\usepackage{hyperref}
\usepackage{arydshln}
\usepackage{bm}
\usepackage{booktabs} 
\usepackage{multirow}
\usepackage{float}
\restylefloat{table}
\usepackage{enumitem}
\usepackage[font={normalsize}]{caption}

\usepackage[ruled, algo2e, noend]{algorithm2e}

\DeclareMathAlphabet{\altmathcal}{OMS}{cmsy}{m}{n}
\usepackage{dblfloatfix}  

\newcommand{\s}[1]{\color{customgray} #1}

\usepackage{tikz}
\definecolor{cell}{HTML}{999999}
\definecolor{lightgray}{HTML}{F3F3F3}
\definecolor{customgray}{HTML}{585858}
\definecolor{rowbackground}{HTML}{F9F9F9}

\usepackage{color, colortbl}

\newcommand{\f}{
\begin{tikzpicture}[every node/.style={inner sep=0,outer sep=0},scale=0.25]
    \fill [rounded corners=0.08cm,fill=cell] (0,0)--(.6,0)--(.6,.6)--(0,.6)--cycle;
\end{tikzpicture}
}

\newcommand{\e}{
\begin{tikzpicture}[every node/.style={inner sep=0,outer sep=0},scale=0.2]
    \fill [rounded corners=0.08cm,fill=lightgray] (0,0)--(.6,0)--(.6,.6)--(0,.6)--cycle;
\end{tikzpicture}
}

\newcommand{\method}{\textsc{UnMask}}

\newcommand{\dataset}{\textsc{\method Dataset}}

\definecolor{pink}{RGB}{255, 105, 180}

\newcommand{\hide}[1]{}

\begin{document}

\title{\method: Adversarial Detection and Defense Through Robust Feature Alignment}


\author{\IEEEauthorblockN{Scott Freitas}
\IEEEauthorblockA{
\textit{Georgia Tech}\\
Atlanta, USA \\
safreita@gatech.edu}
\and
\IEEEauthorblockN{Shang-Tse Chen}
\IEEEauthorblockA{
\textit{National Taiwan University}\\
Taipei, Taiwan \\
stchen@csie.ntu.edu.tw}
\and
\IEEEauthorblockN{Zijie J. Wang}
\IEEEauthorblockA{
\textit{Georgia Tech}\\
Atlanta, USA \\
jayw@gatech.edu}
\and
\IEEEauthorblockN{Duen Horng Chau}
\IEEEauthorblockA{
\textit{Georgia Tech}\\
Atlanta, USA \\
polo@gatech.edu}
}

\maketitle

\begin{abstract}
Recent research has demonstrated that deep learning architectures are vulnerable to adversarial attacks, highlighting the vital need for defensive techniques to detect and mitigate these attacks before they occur. 
We present \method{}, an adversarial detection and defense framework based on robust feature alignment. 
\method{} combats adversarial attacks by extracting robust features (e.g., beak, wings, eyes) from an image (e.g., ``bird'') and comparing them to the expected features of the classification. 
For example, if the extracted features for a ``bird'' image are \textit{wheel}, \textit{saddle} and \textit{frame}, the model may be under attack.
\method{} detects such attacks and defends the model by rectifying the misclassification, re-classifying the image based on its robust features. Our extensive evaluation shows that \method{} \textit{detects} up to 96.75\% of attacks, and \textit{defends} the model by correctly classifying up to 93\% of adversarial images 
produced by the current strongest attack, Projected Gradient Descent,
in the gray-box setting.
\method{} provides significantly better protection than adversarial training across 8 attack vectors, averaging 31.18\% higher accuracy.
We open source the code repository and data with this paper: \url{https://github.com/safreita1/unmask}.

\end{abstract}

\begin{IEEEkeywords}
deep learning, adversarial defense, robust features, adversarial detection
\end{IEEEkeywords}

\section{Introduction}
\begin{figure*}[t!]
    \centering
    \includegraphics[width=0.8\linewidth]{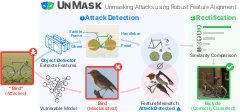}
    \caption{
    \method{} combats adversarial attacks (in red) through extracting \textit{robust features} from an image (``Bicycle'' at top),
    and comparing them to expected features of the classification (``Bird'' at bottom) from the unprotected model.
    Low feature overlap signals an attack.
    \method{} rectifies  misclassification using the image's extracted features.
    Our approach \textit{detects} 96.75\% of gray-box attacks (at 9.66\% false positive rate) and \textit{defends} the model by correctly classifying up to 93\% of adversarial images crafted by Projected Gradient Descent (PGD).
    }
    \label{crown}
\end{figure*}

In the past few years, deep neural networks (DNNs) have shown significant susceptibility to adversarial perturbation \cite{adversarial,originalaa}.
More recently, a wide range of adversarial attacks~\cite{boundaryattack,physical,fgsm} have been developed to defeat deep learning systems---in some cases by changing the value of only a few pixels \cite{singlepixelattack}. 
The ability of these micro perturbations to confuse deep learning architectures highlights a critical issue with modern computer vision systems---that these deep learning systems do not distinguish objects in ways that humans would \cite{inpainting,shapeshop}.
For example, when humans see a bicycle, we see its handlebar, frame, wheels, saddle, and pedals (Fig.~\ref{crown}, top).
Through our visual perception and cognition, 
we synthesize these detection results with our knowledge to determine that we are actually seeing a bicycle. 

However, when an image of a bicycle is adversarially perturbed to fool the model into misclassifying it as a bird (by manipulating pixels), to humans, we still see the bicycle's \textbf{robust features} (e.g., handlebar).
On the other hand, the attacked model fails to perceive these robust features, and is tricked into misclassifying the image. 
How do we incorporate this intuitive detection capability natural to human beings, into deep learning models to protect them from harm?

It was recently posited that adversarial vulnerability is a consequence of a models' sensitivity to well-generalizing features in the data~\cite{ilyas2019adversarial,tsipras2018robustness}. 
Since models are trained to maximize accuracy, they use \textit{any} available information to achieve this goal.
This often results in the use of human incomprehensible features, since a ``head'' or ``wheel'' is as natural to a classifier as any other predictive feature. 
These human incomprehensible (non-robust) features, while useful for improving accuracy, can lead to the creation of adversarially vulnerable models~\cite{ilyas2019adversarial}.
We extend this notion of adversarial vulnerability as a consequence of non-robust features and develop a framework that protects against attacks by incorporating human priors into the classification pipeline.

\subsection{Contributions}
\noindent\textbf{1. Robust Feature Extraction.} 
We contribute the idea that \textit{robust feature alignment} offers a powerful, explainable and practical method of detecting and defending against adversarial perturbations in deep learning models. 
A significant advantage of our proposed concept is that while an attacker may be able to manipulate the class label by subtly changing pixel values, 
it is much more challenging to simultaneously manipulate all the individual features that jointly compose the image. 
We demonstrate that by adapting an object detector, we can effectively extract higher-level \textit{robust features} 
contained in images to detect and defend against adversarial perturbations. 
(Section~\ref{sec:features})

\smallskip
\noindent\textbf{2. \method: Detection \& Defense Framework.} Building on our core concept of robust feature alignment, we propose \method{} as a framework to detect and defeat adversarial attacks by quantifying the similarity between the image's extracted features with the expected features of its predicted class. 
To illustrate how \method{} works, we use the example from Figure~\ref{crown}, where a bicycle image has been attacked such that it would fool an unprotected model into  misclassifying it as a bird.
For a real ``bird'' image, we would expect to see features such as
\textit{beak}, \textit{wing} and \textit{tail}. 
However, \method{} would (correctly) extract bike features: 
\textit{wheel}, \textit{frame}, and \textit{pedals}.
\method{} quantifies the similarity between the \textit{extracted} features (of a bike) with the \textit{expected} features (of a bird), in this case zero. 
This comparison gives us the dual ability to both \textbf{detect} adversarial perturbations by selecting a similarity threshold for which we classify an image as adversarial, and to \textbf{defend} the model by predicting a corrected class that best matches the extracted features. 
(Section~\ref{sec:detect_defend})

\smallskip
\noindent\textbf{3. Extensive Evaluation.} 
We extensively evaluate \method{}'s effectiveness using the large \dataset{} that we have newly curated, with over 21k images in total.
We test multiple factors, including:
4 strong attacks;
2 attack strength levels;
2 popular CNN architectures;
and multiple combinations of varying numbers of classes and feature overlaps.
Experiments demonstrate that our approach \textit{detects} up to 
96.75\% of gray-box attacks with a false positive rate of 9.66\% and (2) \textit{defends} the model by correctly classifying up to 93\% of adversarial images crafted by Projected Gradient Descent (PGD).
\method{} provides significantly better protection than adversarial training across 8 attack vectors, averaging 31.18\% higher accuracy.
(Section~\ref{eval})

\smallskip
\noindent\textbf{4. Reproducible Research: Open-source Code \& Dataset.} 
We contribute a new dataset incorporating PASCAL-Part \cite{pascalparts}, PASCAL VOC 2010 \cite{voc2010}, a subset of ImageNet \cite{deng2009imagenet} and images scraped from Flickr---which we call the \dataset. The goal of this dataset is extend the PASCAL-Part and PASCAL VOC 2010 dataset in two ways---(1) by adding 9,236 and 6,592 manually evaluated images from a subset of ImageNet and Flickr, respectively; and (2) by converting PASCAL-Part to the standard Microsoft COCO format \cite{coco} for easier use and adoption by the research community. 
Furthermore, we release this new dataset along with our code and models on GitHub, at \url{https://github.com/safreita1/unmask}.

Throughout the paper we follow standard notation, using capital bold letters for matrices (e.g., $\bm{A}$), lower-case bold letters for vectors (e.g., $\bm{a}$) and calligraphic font for sets (e.g., $\altmathcal{S}$). 

\section{Background and Related Work}\label{related}
Adversarial attacks typically operate in one of three threat models---(i) white-box, (ii) gray-box or (iii) black-box. In the (i) white-box setting, everything about the model and defense techniques is visible to the attacker, allowing them to tailor attacks to individual neural networks and defense techniques. This is the hardest scenario for the defender since the adversary is aware of every countermeasure. In (ii) the gray-box threat model, we assume that the attacker has access to the classification model but no information on the defense measures. In (iii) the black-box setting, we assume that the attacker has no access to the classification model or the defense techniques. This is the most difficult, and realistic scenario for the attacker since they typically have limited access to the model's inner workings. Despite this disadvantage, recent research has shown that it's possible for adversaries to successfully craft perturbations for deep learning models in the black-box setting \cite{genattack,blackboxattack2,blackboxattack}. 
\subsection{Adversarial Attacks} There exists a large body of adversarial attack research. We provide a brief background on the attacks we use to probe the robustness of the \method{} detection and defense framework. We assume that all attack models operate in a gray-box setting, where the attacker has full knowledge of the classification model, but no knowledge of the defensive measures.
We focus on untargeted attacks in all of the experiments.

\medskip
\noindent
\textbf{Projected Gradient Descent} (PGD)~\cite{madry2017towards} finds an adversarial example $\bm{X}_p$ by iteratively maximizing the loss function $J(\bm{X}_p, y)$ for $T$ iterations, where $J$ is the cross-entropy loss.

\begin{equation}
\label{eq:pgd}
    \bm{X}^{(t+1)}_p = \bm{X}^{(t)}_p + \Pi_{\tau}\Big[\epsilon \cdot sign\Big\{\nabla_{\bm{X}^{(t)}_p} J(\bm{X}^{(t)}_p,y)\Big\}\Big]
\end{equation}

Here $\bm{X}_p^{(0)} = \bm{X}$ and at every step $t$, the previous perturbed input $\bm{X}^{(t-1)}_p$ is modified with the sign of the gradient of the loss, multiplied by $\epsilon$ (attack strength). 
$\Pi_{\tau}$ is a function that clips the input at the positions where it exceeds the predefined $L_\infty$ (or $L_2$) bound $\tau$. We select PGD since it's one of the strongest first order attacks \cite{madry2017towards}.

\medskip
\noindent
\textbf{MI-FGSM} (MIA)~\cite{dong2018boosting} is a gradient-based attack utilizing momentum, where it accumulates the gradients $\bm{g}_t$ of the first $t$ iterations with a decay factor $\mu$. 

\begin{equation}\label{eq:mia}
    \bm{g}^{(t+1)} = \mu \cdot \bm{g}^{(t)} + \frac{\nabla_{\bm{X}^{(t)}_p} J(\bm{X}^{(t)}_p, y)}{||\nabla_{\bm{X}^{(t)}_p} J(\bm{X}^{(t)}_p, y)||_1}
\end{equation}

\begin{equation}\label{eq:mia2}
    \bm{X}^{(t+1)}_p = \bm{X}^{(t)}_p + \Pi_{\tau}\Big[\ \alpha \cdot sign(\bm{g}^{(t+1)}) \Big]
\end{equation}

Here $\alpha=\frac{\epsilon}{T}$, which controls the attack strength. The gradient accumulation (momentum) helps alleviate the trade-off between attack strength and transferability---demonstrated by winning the NIPS 2017 Targeted and Non-Targeted Adversarial Attack competitions.

\subsection{Adversarial Defense \& Detection} 

\medskip
\noindent
\textbf{Adversarial training} seeks to vaccinate deep learning models to adversarial image perturbations by modifying the model's training process to include examples of attacked images~\cite{advtraining1,advtraining2}. 
The idea is that the model will learn to classify these adversarial examples correctly if enough data is seen. It is one of the current state-of-the-art defenses in the white-box setting.  
When the adversarial examples are crafted by PGD, it is known to improve robustness even against other types of attacks, because PGD is the strongest first-order attack and approximately finds the hardest examples to train~\cite{madry2017towards}. 
The adversarial training process can be seen in the equation~\ref{eq:l-adv} below, where we utilize the adversarial perturbations generated by equation~\ref{eq:pgd}.

\begin{equation}\label{eq:l-adv}
    \underset{\bm{W}}{\text{min }} \Big[ \mathbb{E}_{(\bm{X},y) \sim D} \Big(\underset{\delta\in \altmathcal{S}}{\text{max}}\; L(\bm{W}, \bm{X} + \delta, y)\Big)\Big] 
\end{equation}

The downside to this technique is that it requires a large amount of data and training time; in addition, it does not incorporate a human prior into the training process.

\medskip
\noindent
\textbf{Alternative Defenses.} Defensive distillation is one technique used to robustify deep learning models to adversarial perturbation by training two models---one where the model is trained normally using the provided hard labels and a second model which is trained on soft labels from the probability output of the first model~\cite{distill1}. However, it has been show that type of technique is likely a kind of gradient masking, making it vulnerable to black-box transfer attacks~\cite{distillattack}. In addition, there are many other defensive techniques, some of which include pre-processing the data---which has the goal of eliminating adversarial perturbation before model sees it. A couple of proposed techniques include, image compression~\cite{shield} and dimensionality reduction~\cite{pcadefense}. Data pre-processing defense is usually model independent and can easily be used along side with other defenses.
The downside of this approach is that most pre-processing techniques have no knowledge of whether the system is actually being attacked. More advanced attacks have also been proposed by replacing the non-differentiable pre-processing step with an approximate differentiable function and back-propogating through the whole pipeline~\cite{shin2017jpeg,athalye2018obfuscated}.

\medskip
\noindent
\textbf{Adversarial detection} attempts to determine whether or not an input is benign or adversarial. This has been studied from multiple perspectives using a variety of techniques, from topological subgraph analysis~\cite{subgraphdetection} to image processing~\cite{imageprocessing,mutation,Xu0Q18}, and hidden/output layer activation distribution information~\cite{distribution1,distribution2,meng2017magnet}. 

\section{\method: Detection and Defense Framework}
\label{methodology} 

\method{} is a new method for protecting deep learning models from adversarial attacks by 
using \textit{robust features} that semantically align with human intuition to combat adversarial perturbations (Figure~\ref{crown}).
The objective is to defend a \textbf{vulnerable} deep learning model \bm{$M$} (Figure~\ref{crown}, bottom) using our \textbf{\method{}} defense framework \bm{$D$} (Figure~\ref{crown}, top),
where the adversary has full access to $M$ but is unaware of the defense strategy $D$,
constituting a \textit{gray-box} attack on the overall classification pipeline~\cite{shield}.
In developing the \method{} framework for adversarial detection and defense, we address three sub-problems:

\smallskip
\begin{enumerate}[leftmargin=*,label=\arabic*.]
    \item \textbf{Identify robust features} by training a model $K$ that takes as input, input space $\bm{X}$ and maps it to a set of robust features $\altmathcal{F}=\{K: \bm{X} \rightarrow \mathbb{R}\}$.
    
    \item \textbf{Detect} if input space $\bm{X}$ is benign (+1) or adversarially perturbed (-1) by creating a binary classifier $C:\altmathcal{F} \rightarrow \{\pm 1\}$ that utilizes robust features $\altmathcal{F}$.
    
    \item \textbf{Defend} against adversarial attacks by developing a classifier $C: \altmathcal{F} \rightarrow y$ to predict $y$ using robust features $\altmathcal{F}$.
\end{enumerate}

\smallskip
We present our solution to problem 1 in
Section~\ref{sec:features}; and discuss how to solve problems 2 and 3 in Section~\ref{sec:detect_defend}.

\subsection{Aligning Robust Features with Human Intuition}
\label{sec:features}
In order to discuss adversarially robust features, we categorize features into three distinct types: \textbf{$p$-useful,} (ii) \textbf{$\gamma$-robust}, and (iii) \textbf{useful but non-robust}~\cite{ilyas2019adversarial}. 
For a distribution $D$, a feature is defined as $p$-useful ($p>0$) if it is correlated with the true label in expectation (Eq.~\ref{eq:p-uesful}). These $p$-useful features are important for obtaining high accuracy classifiers.

\begin{equation} \label{eq:p-uesful}
    \underset{(\bm{X},y) \sim D}{\mathbb{E}}[y \cdot f(\bm{X})] \geq p
\end{equation}

A feature is referred to as a \textit{robust feature} ($\gamma$-robust) if that feature remains useful ($\gamma > 0$) under a set of allowable adversarial perturbations $\delta \in S$. This is defined in Equation~\ref{eq:g-uesful}.

\begin{equation} \label{eq:g-uesful}
    \underset{(\bm{X},y) \sim D}{\mathbb{E}}\left[\underset{\delta \in S}{\text{inf}}\, y \cdot f(\bm{X}+ \delta)\right] \geq \gamma
\end{equation}

Lastly, a useful but non-robust feature is a feature that is $p$-useful for some $p>0$ but not $\gamma$-robust for any $\gamma \geq 0$. These non-robust features are useful for obtaining high accuracy classifiers, but can be detrimental in the presence of an adversary since they are often weakly correlated with the label and can be easily perturbed.

When training a classifier, minimizing classification loss makes no distinction between robust and non-robust features---it only distinguishes whether a feature is $p$-useful. This causes a model to utilize useful but non-robust features in order to improve classifier accuracy. However, in the presence of adversarial perturbation, these useful but non-robust features become anti-correlated with the label and lead to misclassification. On the other hand, if a model is trained from purely \textit{robust features}, it has lower classification performance; but gains non-trivial adversarial robustness \cite{ilyas2019adversarial}. 

\begin{table}[tb]
\centering
\small
\setlength{\tabcolsep}{1.5pt}
\renewcommand{\arraystretch}{1}

\begin{tabular}{@{}l
llllllllllllllllllll@{}
}

\bfseries Features & \rotatebox{90}{Airplane} & \rotatebox{90}{Bicycle} & \rotatebox{90}{Bird} & \rotatebox{90}{Boat} & \rotatebox{90}{Bottle} & \rotatebox{90}{Bus} & \rotatebox{90}{Car} & \rotatebox{90}{Cat} & \rotatebox{90}{Chair} & \rotatebox{90}{Cow} & \rotatebox{90}{Dining Table} & \rotatebox{90}{Dog} & \rotatebox{90}{Horse} & \rotatebox{90}{Motorbike} & \rotatebox{90}{Person} & \rotatebox{90}{Potted Plant} & \rotatebox{90}{Sheep} & \rotatebox{90}{Sofa} & \rotatebox{90}{Train} & \rotatebox{90}{Television} \\

\midrule

Arm 
& \e & \e & \e & \e & \e & \e & \e & \e & \e & \e & \e & \e & \e & \e & \f & \e & \e & \e & \e & \e \\

\rowcolor{rowbackground} 
Beak
& \e & \e & \f & \e & \e & \e & \e & \e & \e & \e & \e & \e & \e & \e & \e & \e & \e & \e & \e & \e \\

Body 
& \f & \e & \e & \e & \f & \e & \e & \e & \e & \e & \e & \e & \e & \e & \e & \e & \e & \e & \e & \e \\
 
\rowcolor{rowbackground} 
Cap 
& \e & \e & \e & \e & \f & \e & \e & \e & \e & \e & \e & \e & \e & \e & \e & \e & \e & \e & \e & \e \\
 
\mdseries Coach
& \e & \e & \e & \e & \e & \e & \e & \e & \e & \e & \e & \e & \e & \e & \e & \e & \e & \e & \f & \e \\

\rowcolor{rowbackground} 
Door
& \e & \e & \e & \e & \e & \f & \f & \e & \e & \e & \e & \e & \e & \e & \e & \e & \e & \e & \e & \e \\
 
Engine
& \f & \e & \e & \e & \e & \e & \e & \e & \e & \e & \e & \e & \e & \e & \e & \e & \e & \e & \e & \e \\

\rowcolor{rowbackground} 
Ear
& \e & \e & \e & \e & \e & \e & \e & \f & \e & \f & \e & \f & \f & \e & \f & \e & \f & \e & \e & \e \\

Eye
& \e & \e & \f & \e & \e & \e & \e & \f & \e & \f & \e & \f & \f & \e & \f & \e & \f & \e & \e & \e \\

\rowcolor{rowbackground} 
Eyebrow 
& \e & \e & \e & \e & \e & \e & \e & \e & \e & \e & \e & \e & \e & \e & \f & \e & \e & \e & \e & \e \\

Foot
& \e & \e & \f & \e & \e & \e & \e & \e & \e & \e & \e & \e & \e & \e & \f & \e & \e & \e & \e & \e \\
 
\rowcolor{rowbackground} 
Front side 
& \e & \e & \e & \e & \e & \f & \f & \e & \e & \e & \e & \e & \e & \e & \e & \e & \e & \e & \e & \e \\

Hair 
& \e & \e & \e & \e & \e & \e & \e & \e & \e & \e & \e & \e & \e & \e & \f & \e & \e & \e & \e & \e \\

\rowcolor{rowbackground} 
Hand 
& \e & \e & \e & \e & \e & \e & \e & \e & \e & \e & \e & \e & \e & \e & \f & \e & \e & \e & \e & \e \\
 

Head 
& \e & \e & \f & \e & \e & \e & \e & \f & \e & \f & \e & \f & \f & \e & \f & \e & \f & \e & \f & \e \\
 
\rowcolor{rowbackground} 
Headlight
& \e & \e & \e & \e & \e & \f & \f & \e & \e & \e & \e & \e & \e & \f & \e & \e & \e & \e & \f & \e \\

Hoof
& \e & \e & \e & \e & \e & \e & \e & \e & \e & \e & \e & \e & \f & \e & \e & \e & \e & \e & \e & \e \\

\rowcolor{rowbackground}
Horn
& \e & \e & \e & \e & \e & \e & \e & \e & \e & \f & \e & \e & \e & \e & \e & \e & \f & \e & \e & \e \\

Leg
& \e & \e & \f & \e & \e & \e & \e & \f & \e & \f & \e & \e & \f & \e & \f & \e & \f & \e & \e & \e \\
 
\rowcolor{rowbackground} 
License plate
& \e & \e & \e & \e & \e & \f & \f & \e & \e & \e & \e & \e & \e & \e & \e & \e & \e & \e & \e & \e \\
 
Mirror
& \e & \e & \e & \e & \e & \f & \f & \e & \e & \e & \e & \e & \e & \e & \e & \e & \e & \e & \e & \e \\

\rowcolor{rowbackground} 
Mouth
& \e & \e & \e & \e & \e & \e & \e & \e & \e & \e & \e & \e & \e & \e & \f & \e & \e & \e & \e & \e \\

Muzzle
& \e & \e & \e & \e & \e & \e & \e & \e & \e & \f & \e & \f & \f & \e & \e & \e & \f & \e & \e & \e \\

\rowcolor{rowbackground} 
Neck
& \e & \e & \f & \e & \e & \e & \e & \f & \e & \f & \e & \f & \f & \e & \f & \e & \f & \e & \e & \e \\

Nose
& \e & \e & \e & \e & \e & \e & \e & \f & \e & \e & \e & \f & \e & \e & \f & \e & \e & \e & \e & \e \\

\rowcolor{rowbackground} 
Paw
& \e & \e & \e & \e & \e & \e & \e & \f & \e & \e & \e & \f & \e & \e & \e & \e & \e & \e & \e & \e \\

Plant
& \e & \e & \e & \e & \e & \e & \e & \e & \e & \e & \e & \e & \e & \e & \e & \f & \e & \e & \e & \e \\

\rowcolor{rowbackground} 
Pot
& \e & \e & \e & \e & \e & \e & \e & \e & \e & \e & \e & \e & \e & \e & \e & \f & \e & \e & \e & \e \\

Saddle
& \e & \f & \e & \e & \e & \e & \e & \e & \e & \e & \e & \e & \e & \f & \e & \e & \e & \e & \e & \e \\

\rowcolor{rowbackground} 
Screen
& \e & \e & \e & \e & \e & \e & \e & \e & \e & \e & \e & \e & \e & \e & \e & \e & \e & \e & \e & \f \\

Stern
& \f & \e & \e & \e & \e & \e & \e & \e & \e & \e & \e & \e & \e & \e & \e & \e & \e & \e & \e & \e \\

\rowcolor{rowbackground} 
Tail 
& \f & \e & \f & \e & \e & \e & \e & \f & \e & \f & \e & \f & \f & \e & \e & \e & \f & \e & \e & \e \\

Torso
& \e & \e & \f & \e & \e & \e & \e & \f & \e & \f & \e & \f & \f & \e & \f & \e & \f & \e & \e & \e \\

\rowcolor{rowbackground} 
Vehicle
& \e & \e & \e & \e & \e & \f & \f & \e & \e & \e & \e & \e & \e & \e & \e & \e & \e & \e & \e & \e \\

Wheel 
& \f & \f & \e & \e & \e & \f & \f & \e & \e & \e & \e & \e & \e & \f & \e & \e & \e & \e & \e & \e \\
 
\rowcolor{rowbackground} 
Window
& \e & \e & \e & \e & \e & \f & \f & \e & \e & \e & \e & \e & \e & \e & \e & \e & \e & \e & \e & \e \\

Wing
& \f & \e & \f & \e & \e & \e & \e & \e & \e & \e & \e & \e & \e & \e & \e & \e & \e & \e & \e & \e \\

\noalign{\vskip 1mm}    

\textbf{Class Set }
&   \multicolumn{20}{c}{} \\

\midrule

CS3a
& \e & \e & \e & \e & \e & \e & \f & \e & \e & \e & \e & \e & \e & \e & \f & \e & \e & \e & \f & \e \\
 
\rowcolor{rowbackground} 
CS3b
& \e & \e & \f & \e & \e & \e & \e & \e & \e & \e & \e & \f & \e & \e & \f & \e & \e & \e & \e & \e \\
 
CS5a
& \e & \e & \e & \e & \f & \e & \f & \e & \e & \e & \e & \f & \e & \e & \f & \e & \e & \e & \f & \e \\

\rowcolor{rowbackground} 
CS5b
& \e & \e & \f & \e & \e & \e & \f & \e & \e & \e & \e & \f & \e & \e & \f & \e & \e & \e & \f & \e \\

\addlinespace
\end{tabular}
\caption{Class-Feature Matrix. 
Top: dots mark classes' features.
Bottom: four class sets with varying levels of feature overlap. Features \textit{vehicle} and \textit{coach} have sub-features not listed here due to space (see Github repository). 
}

\label{feature_matrix}
\end{table}

Our goal is to enable a model $M$ to use both $\gamma$-robust and useful but non-robust features to achieve the highest possible classification accuracy, while utilizing only the $\gamma$-robust features to determine if an image is attacked. 
This allows model $M$ to use \textit{any} signal in the data to improve classification accuracy, while the defense framework uses \textit{only} the robust features to provide a safeguard against adversarial perturbations.
In order to determine the $\gamma$-robust features from an input space $\bm{X}$, we develop model $K$ to identify robust features by training on a robust distribution $\hat{D}_R$ that satisfies Equation~\ref{eq:robust-dist}.

\begin{equation} \label{eq:robust-dist}
    \underset{(\bm{X},y) \sim \hat{D}_R}{\mathbb{E}} [f(\bm{X}) \cdot y] =     \begin{cases}
        \underset{(\bm{X},y) \sim D}{\mathbb{E}} [f(\bm{X}) \cdot y] & \text{if}\, f \in \altmathcal{F} \\
        0,              & \text{otherwise}
    \end{cases}
\end{equation}

Here $\altmathcal{F}$ represents the set of human-level robust features identified in the Pascal-Part dataset using segmentation masks. Ideally, we want these human-level features to be as useful as the original distribution $D$, while excluding the useful but non-robust features.
Using robust dataset $\hat{D}_R$, we train a robust feature extraction model $K$ with weights $\bm{W}$ to recognize \textit{only} the human-level robust features using Equation~\ref{eq:unmask-training}.

\begin{equation}\label{eq:unmask-training}
    \underset{\bm{W}}{\text{min}}\;\left[ \mathbb{E}_{(\bm{X},y) \sim \hat{D}_R} L(\bm{X}, y)\right]
\end{equation}

From a practical standpoint, we adopt a Mask R-CNN architecture~\cite{maskrcnn} for feature extraction model $K$. 
We considered multiple approaches, 
but decided to use Mask R-CNN
for its ability to leverage image segmentation masks to learn and identify coherent image regions that closely resemble robust features that would appear semantically and visually meaningful to humans. 
Different from conventional image classification models or object detectors,  
the annotations used to train our robust feature extractor $K$ are \textit{segmented} object parts instead of the whole objects.
For example, for the \textit{wheel} feature, an instance of training data would consist of a bike image and a \textit{segmentation mask} indicating which region of that image represents a wheel.  
Technically, this means $K$ uses only a part of an image, and not the whole image, for training (See Table~\ref{feature_matrix}).
Furthermore, while an image may consist of multiple image parts, $K$ treats them independently.

\subsection{Robust Features For Detection and Defense} 
\label{sec:detect_defend}

Leveraging the robust features extracted from model $K$, we introduce \method{} as a detection and defense framework ($D$). 
For an \textit{unprotected} model $M$ (Figure~\ref{crown}, bottom),
an adversary crafts an \textit{attacked} image by carefully manipulating its pixel values using an adversarial technique (e.g., PGD~\cite{madry2017towards}).
This attacked image then fools model $M$ into misclassifying the image,
as shown in Figure~\ref{crown}.
To guard against this kind of attack on $M$, we use our \method{} defense framework $D$
in conjunction with the \textit{robust feature extraction model} $K$ (Figure~\ref{crown}, top). 

Model $K$ processes the same image, which may be benign or attacked, and
extracts the robust features from the image to compare to the images' expected features.
Figure~\ref{crown} shows an example, where an attacked \textit{bike} image has fooled the unprotected model $M$ to classify it as a \textit{bird}. We would \textit{expect} the robust features to include \textit{head}, \textit{claw}, \textit{wing}, and \textit{tail}.
However, from the same (attacked) image, \method{}'s model $K$ extracts \textit{wheels}, \textit{handle} and \textit{seat}.
Comparing the set of \textit{expected} features and the actual \textit{extracted} features (which do not overlap in this example), \method{} determines the image was attacked, and predicts its class to be \textit{bike} based on the extracted features.
This robust feature alignment forges a layer of protection around a model by disrupting the traditional pixel-centric attack \cite{boundaryattack,madry2017towards,singlepixelattack}. This forces the adversary to solve a more complex problem of manipulating both the class label and all of the image's constituent parts. For example, in Figure~\ref{crown} the attacker needs to fool the defensive layer into misclassifying the bike as a bird by, (1) changing the class label and (2) manipulating the robust bike features (\textit{wheel}, \textit{seat}, \textit{handlebar}) into bird features. 
\method{}'s  technical operations for detection and defense are detailed below and in Algorithm \ref{unmask_alg}:

\begin{enumerate}[label=\arabic*., topsep=4pt, leftmargin=*, itemsep=3pt]
    \item \textbf{Classify} input space $\bm{X}$ to obtain prediction 
    $\hat{y}$ from unprotected model $M$, i.e.,
    $\hat{y} = M(\bm{X})$.
    At this point, \method{} does not know if $\bm{X}$ is adversarial or not.
    
    \item \textbf{Extract robust features} of $\bm{X}$ using robust feature extraction model $K$, i.e.,
    $f_r = K(\bm{X})$, where $f_r \subseteq \altmathcal{F}$.
    Armed with these features $f_r$, \method{} detects if model $M$ is under attack, and rectifies misclassification.
    
    \item \textbf{Detect attack} by measuring the similarity between the \textit{extracted} features $f_r$ and the set of \textit{expected} features $f_e = \bm{V}[\hat{y}]$---where $\bm{V}$ is the feature matrix in Table~\ref{feature_matrix}---by calculating the Jaccard similarity score  $s = JS(f_e, f_a)$.
    If distance score $d = 1 - s$ is greater than threshold $t$,
    input $\bm{X}$ is deemed benign, otherwise adversarial.
    Adjusting $t$ allows us to assess the trade-off between sensitivity and specificity, which we describe in detail in Section~\ref{eval}.
    
    \item \textbf{Defend and rectify} an input to be adversarial also means that model $M$ is under attack and is giving unreliable classification output.
    Thus, we need to rectify the misclassification.
    \method{} accomplishes this by comparing the robust extracted features $f_r$ to every set of class features in $\bm{V}$, outputting class $\hat{y}$ that contains the highest feature similarity score $s$, where $0 \leq s \leq 1$. 
\end{enumerate}

\begin{algorithm2e}[t]
    \KwIn{Data distribution $D$, unprotected model $M$, class-feature matrix $\bm{V}$, input space $\bm{X}_t$, threshold $t$}
    \KwResult{adversarial prediction $z \in \{-1, 1\}$, predicted class $p$}
\BlankLine
    
    $\underset{(\bm{X},y) \sim \hat{D}_R}{\mathbb{E}} [f(\bm{X}) \cdot y] =     
        \begin{cases}
        \underset{(\bm{X},y) \sim D}{\mathbb{E}} [f(\bm{X}) \cdot y] & \text{if}\, f \in \altmathcal{F} \\
        0,              & \text{otherwise}
        \end{cases}$ \hfill (create $\hat{D}_R$)
        
    \BlankLine
    
    $K = \underset{\bm{W}}{\text{min}}\;\left[ \mathbb{E}_{(\bm{X},y) \sim \hat{D}_R} L(\bm{X}, y)\right]$; \\

    \BlankLine
    
    $f_r$ = $K(\bm{X}_t)$; \hfill (extracted features)\\
    \BlankLine
    
    $f_e$ = $\bm{V}[M(\bm{X}_t)]$; \hfill (expected features)\\
    
\BlankLine
    \textbf{\textit{Detection:}}\\
    \Indp 
    $s = JS(f_r, f_e);\;d = 1 - s$; \hfill (JS = Jaccard similarity)
    \Indm

    \[
    z = \left\{ \begin{array}{ll} +1\; \text{(benign)} ,  & \,\text{if  } d < t \\
                                  -1\; \text{(adversarial)}, & \,\text{if }  d \ge t
  \end{array} \right.
   \]

    \textbf{\textit{Defense:}}\\
    \[
    p = \left\{ \begin{array}{ll} \hat{y},  & \,\text{if  } z = +1 \\
                                  \underset{c\in C}{\arg\!\min}\; JS(f_e, \bm{V}[c]), & \,\text{if }  z = -1
  \end{array} \right.
   \]

    \Return $z$, $p$\;
        
    \caption{\method}
    \label{unmask_alg}
\end{algorithm2e}

\section{Evaluation}\label{eval}

\begin{table*}[t]
\centering
\setlength{\tabcolsep}{4pt}
\begin{tabular}{lrrrr|rrrrrrrrrrrr} 
\toprule
\multicolumn{5}{c|}{\bfseries Experimental Setup} & 
\multicolumn{3}{c}{\bfseries PASCAL-Part} & 
\multicolumn{1}{c}{\bfseries VOC+Net} &
\multicolumn{2}{c}{\bfseries Flickr} \\

\cmidrule(l){1-5} \cmidrule(l){6-8} \cmidrule(l){9-9} \cmidrule(l){10-11}

Model & Class set & Classes & Parts & Overlap & Train & Val & Test & Train & Val & Test \\
\midrule
\addlinespace[2ex]
\multirow{1}{*}{\textbf{K}} 
& - & 44 & - & - & 7,457 & 930 & 936 & - & - & - \\
\addlinespace[2ex]
\multirow{4}{*}{\textbf{M}} 
& CS3a & 3 & 29 & 6.89\% & - & - & - & 7,780 & 1,099 & 2,351 \\
& CS3b & 3 & 18 & 50.00\% & - & - & - & 9,599 & 1,339 & 2,867 \\
& CS5a & 5 & 34 & 23.53\% & - & - & - & 11,639 & 1,477 & 3,179 \\
& CS5b & 5 & 34 & 29.41\% & - & - & - & 13,011 & 1,928 & 4,129 \\

\bottomrule
\addlinespace
\end{tabular}
\caption{
Number of images used to train and evaluate models $K$, $M$ and defense framework $D$. 
We train $K$ on PASCAL-Part dataset, and model $M$ on PASCAL VOC 2010 plus a subset of ImageNet.
Four \textit{class sets} are investigated in the evaluation, with varying classes and feature overlap.
We evaluate model $M$ and defense framework $D$ on Flickr.}
\label{table:unmask_dataset}
\end{table*}

We extensively evaluate \method{}'s effectiveness in \textbf{defending} and \textbf{detecting} adversarial perturbations using: $4$ strong attacks across two strength levels;
$2$ popular CNN architectures
as unprotected models $M$;
multiple combinations of varying numbers of classes and feature overlaps; and
benchmarking \method{} against one of the strongest adversarial defenses---adversarial training~\cite{madry2017towards}.
All experiments are conducted in a Linux environment on a DGX-1 running Python 3.6 with open-source libraries Keras, Tensorflow, PyTorch, Advertorch~\cite{ding2019advertorch} and Matterport~\cite{matterport};

\subsection{Experiment Setup}\label{detailed_setup}
Experiments are conducted in a Linux environment using Python 3 on an Nvidia DGX-1. 
We open-source all of the code, data and models used in this paper. 
For additional information on using \method{}, we provide a detailed walk through at \url{https://github.com/safreita1/unmask}.

\smallskip
\noindent\textbf{UnMask dataset.}
We curated the \dataset{} for evaluation, which consists of four component datasets---PASCAL-Part, PASCAL VOC 2010, a subset of ImageNet and images scraped from Flickr (Table~\ref{table:unmask_dataset}). 
To ensure the Flickr images are not duplicates, we compare the perceptual hash of each Flickr image to ImageNet and VOC'10. 
The goal of this curation is to 
(i) collect all the data used in our evaluation as a single source for use by the research community, and 
(ii) to increase the number of images available for evaluating the performance of the deep learning models and the \method{} defense framework. 
We designed multiple \textit{class sets} with varying number of classes and feature overlap (e.g., CS3a, in Table~\ref{table:unmask_results}; and Table~\ref{feature_matrix}),
to study how they affect detection and defense effectiveness.
We call each combination of \textit{class count} and \textit{feature overlap} a ``class set'', abbreviated as ``CS.'' 
CS3 thus means a class set with 3 classes.
CS3a and CS3b have the same number of classes, with different feature overlap. 
We further discuss the utilization of data below.

In Table~\ref{feature_matrix}, the class-feature matrix describes the features contained by each class in the dataset.
The PASCAL-Part dataset has 18 variations of the leg feature, however, in order to create a model that better generalizes, we combine this to a single leg feature. 
We note that in Table~\ref{feature_matrix}, that two features have multiple sub-features condensed into a single feature (not listed due to space constraints). 
These features are: vehicle: \{vehicle left, vehicle right, vehicle top, vehicle back\} and coach: \{coach left, coach right, coach back, coach top, coach front\}.
In addition, we note that there is a minor error in the conversion of the handlebar feature in the bike and motorcycle class (handlebar features were labeled as hand). 
However, since those classes are not utilized in the experiments, the effects are minimized.

\begin{figure}[!b]
\centering
\includegraphics[width=0.9\linewidth]{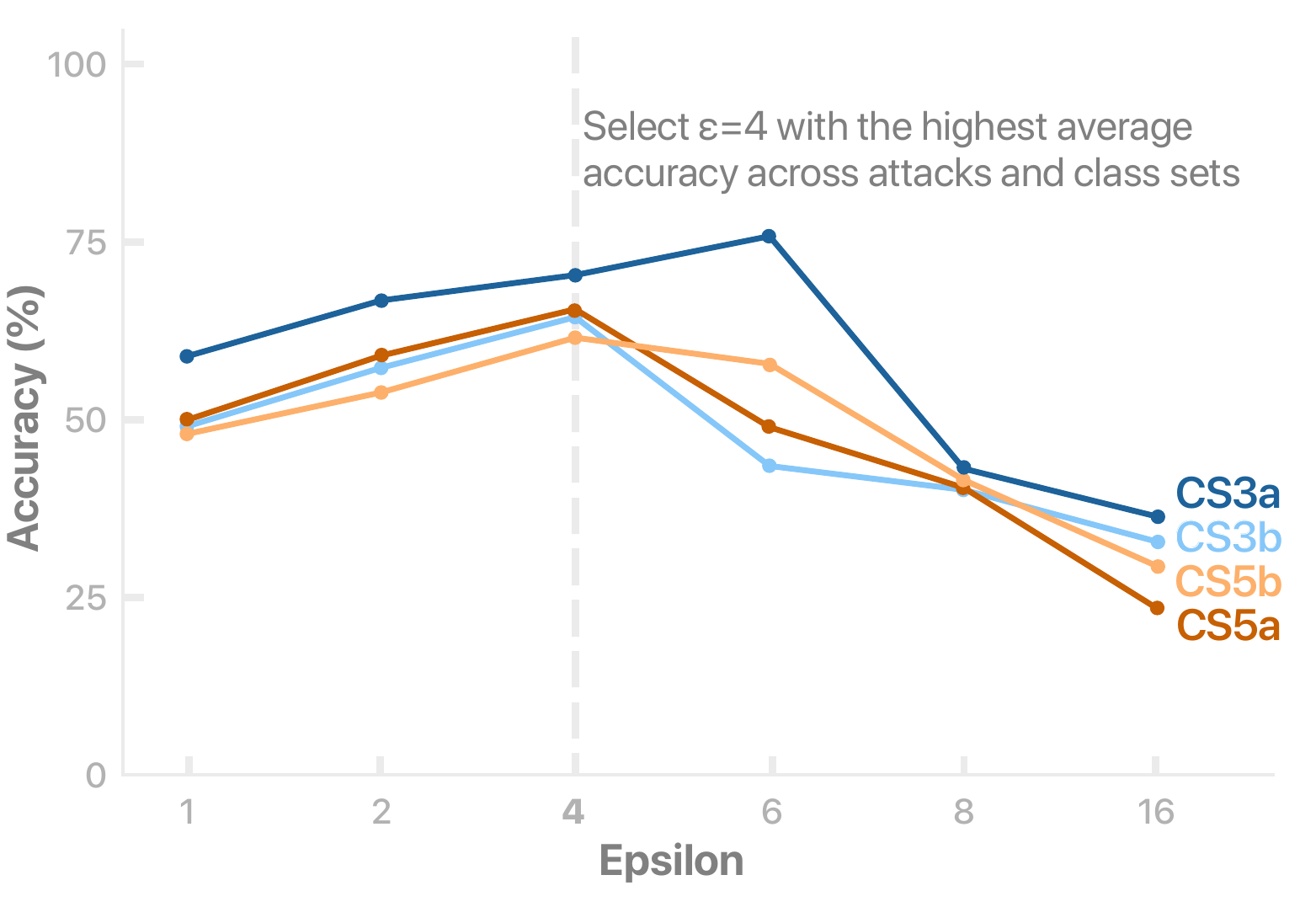}
\caption{Line search for adversarial training parameter $\epsilon$ on validation data. We select $\epsilon$ = $4$, since it provides the best performance on most attacks.}
\label{fig:line_search}
\end{figure}

\smallskip
\noindent\textbf{Adversarial attacks.}
We evaluate \method{} on $4$ attacks: 

\begin{itemize}[itemsep=1ex,leftmargin=*]
    \item \textbf{PGD-$L_\infty$}, one of the strongest first-order attacks \cite{madry2017towards}. 
    Its key parameter $\epsilon$ represents the allowed per-pixel perturbation.
    For example, $\epsilon$ = $4$ means 
    changing up to $4$ units of intensity (out of $255$). 
    It is common to evaluate $\epsilon$ up to $16$, with a stepsize of $2$ and $20$ iterations~\cite{shield,advtraining1,madry2017towards}.
    
    \item \textbf{PGD-$L_2$}
    we also evaluate PGD in the $L_2$ norm, which bounds the $\epsilon$ perturbation across the whole image, instead of a per-pixel bound as in the $L_\infty$ norm.
    Since the perturbation is bounded across the whole image, $\epsilon$ is naturally larger---typically ranging from $300$-$900$~\cite{turner2018clean}.
    
    \item \textbf{MI-FGSM $L_\infty$} (MIA-$L_\infty$)
    is a strong gradient attack with key parameters $\mu$ and $\epsilon$ (see PGD-$L_\infty$). 
    We set decay factor $\mu$=1 as it provides the most effective attack~\cite{dong2018boosting}.
    
    \item \textbf{MI-FGSM $L_2$} (MIA-$L_2$)
    we also evaluate MI-FGSM in the $L_2$ norm, bounding $\epsilon$ perturbation across the whole image, instead of per-pixel as in $L_\infty$ (see PGD-$L_2$).
\end{itemize}

\begin{figure*}[!t]
\centering
\includegraphics[width=0.75\textwidth]{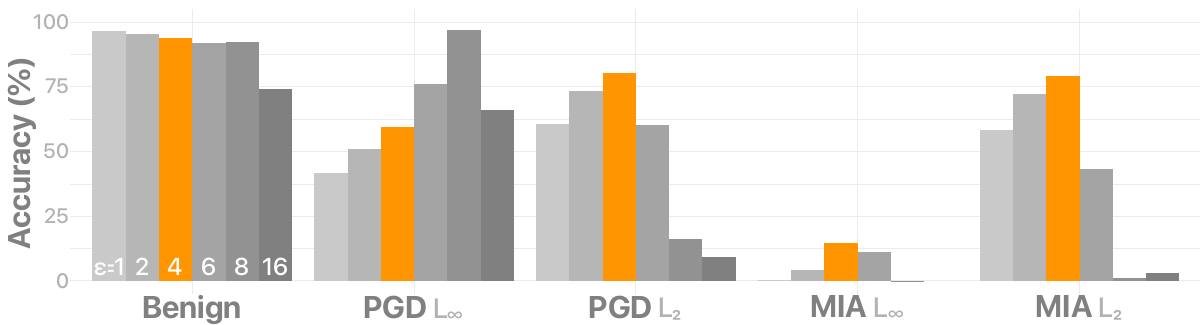}
\caption{Detailed bar chart describing the performance of $\epsilon$ across all attack vectors. We select $\epsilon$ = $4$, since it provides the best performance on most attacks.}
\label{fig:line_search_detailed}
\end{figure*}

\smallskip
\noindent\textbf{Adversarial defense.}
We compare against adversarial training, one of the strongest adversarial defense techniques.
To select $\epsilon$---which controls the perturbation strength of adversarial training---we follow standard procedure~\cite{madry2017towards} and determine $\epsilon$ on a per-dataset basis (class set).
Correctly setting this parameter is critical since a small $\epsilon$ value will have no effect on robustness, while too high a value will lead to poor benign accuracy. 
Following standard procedure, we select $\epsilon$ on a per-dataset basis (class set in our case)~\cite{madry2017towards} by conducting a line search across $\epsilon$ = $\{1, 2, 4, 6, 8, 16\}$.
We find that $\epsilon$ = $4$, provides the best performance on the validation set across each class sets, as seen in Figure~\ref{fig:line_search}. 
While $\epsilon$ = $6$ appears to be a good choice for class set CS3a, it has poor generalization performance and overfits to PGD-$L_\infty$. 
This can be seen through the bar chart of Figure~\ref{fig:line_search_detailed}, where $\epsilon > 4$ reduces the generalization of adversarial training to all other attack vectors.
For this reason, we select $\epsilon$ = $4$ for all class sets.

\smallskip
\noindent\textbf{Training robust feature extraction model.}
As illustrated in Figure~\ref{crown}, the robust feature extraction model $K$ takes an image as input (e.g., bike) and outputs a set of features (e.g., wheel,...).
To train $K$, we use the PASCAL-Part dataset \cite{pascalparts}, which consists of 180,423 feature segmentation masks over 9,323 images across the 44 robust features.
The original dataset contains very fine-grained features, such as 18 types of ``legs'' (e.g., right front lower leg, left back upper leg), 
while for our purposes we only need the abstraction of ``leg''. 
Therefore, we combined these fine-grained features into more generalized ones (shown as rows in Table~\ref{feature_matrix}).

\begin{table}[b]
\centering
\normalsize
\setlength{\tabcolsep}{6pt}
\begin{tabular}{lrrrrrrrrrrrrrrrr} 
\toprule
\multicolumn{1}{c}{\bfseries Class Set} & 
\multicolumn{4}{c}{\bfseries Number of Images Evaluated} \\

\cmidrule(l){1-1} \cmidrule(l){2-5} 

& PGD-$L_\infty$ & PGD-$L_2$ & MIA-$L_\infty$ & MIA-$L_2$ \\
\midrule

CS3a & 3,648 & 4,702 & 4,702 & 4,702 \\
CS3b & 4,652 & 5,732 & 5,734 & 5,734 \\
CS5a & 5,412 & 6,358 & 6,358 & 6,358 \\
CS5b & 6,822 & 8,256 & 8,258 & 8,258 \\

\bottomrule
\addlinespace
\end{tabular}
\caption{
Number of images used to evaluate the detection capability of \method{}. Only images that are successfully attacked are used for evaluation (combined with their benign counterparts), thus the variations in numbers.
We report values for PGD and MIA with $\epsilon$=16/600, respectively. Numbers are similar for $\epsilon$=8/300.}
\label{table:unmask_detection}
\end{table}

We train $K$ for 40 epochs, following a similar procedure described in \cite{matterport}.
We use a ratio of 80/10/10 for training, validating and testing the model respectively (see Table~\ref{table:unmask_dataset}).
Our work is the first adaptation of Mask R-CNN model for the PASCAL-Part dataset.
As such, there are no prior results for comparison.
We computed model $K$'s mean Average Precision (mAP),
which estimates $K$'s ability to extract features. 
The model attains an mAP of 0.56, in line with Mask R-CNN on other datasets \cite{maskrcnn}.
Model $K$ processes up to 4 images per second with a single Nvidia Titan X, matching the speeds reported in \cite{matterport}.
This speed can be easily raised through parallelism by using more GPUs.
As robust feature extraction is the most time-intensive process of the \method{} framework, its speed is representative of the overall  speed of the framework.

\smallskip
\noindent\textbf{Training the unprotected model.}
As described in Section~\ref{methodology},
$M$ is the model under attack, 
and is what \method{} aims to protect.
In practice, the choice of architecture for $M$ and the data it is trained on are determined by the application.
Here, our evaluation studies two popular deep learning architectures --- ResNet50~\cite{resnet} and DenseNet121~\cite{huang2017densely}---however, \method{} supports other architectures. 
Training these models from scratch is generally computationally expensive and requires large amount of data.
To reduce such need for computation and data, 
we adopt the approach described in \cite{matterport}, where we leverage a model pre-trained on ImageNet images, 
and \textit{replace} its dense layers (i.e., the fully connected layers) to enable us to work with various class sets (e.g., CS3a).
Refer to Table \ref{table:unmask_dataset}, for a breakdown of the data used for training and evaluation.

\subsection{Evaluating UnMask Defense and Detection}
\label{analyze}

The key research questions that our evaluation aims to address is how effective \method{} can
(1) \textbf{detect} adversarial images, and
(2) \textbf{defend} against attacks by rectifying misclassification through inferring the actual class label.
We scrape Flickr (see Table~\ref{table:unmask_dataset}) to obtain a large number of unseen images matching our class sets.
We note that evaluation is focused on images containing a single-class (i.e., no ``person'' and ``car'' in same image) as this allows for a more controlled environment.

\begin{figure*}[!b]
\centering
\includegraphics[width=0.8\textwidth]{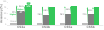}
\caption{Accuracies (in \%) for each class set averaged across all attack vectors, strengths, and models from Table~\ref{table:unmask_results}. 
On average, \method{} (UM) performs 31.18\% better than adversarial training (AT) and 74.44\% than no defense (None).
}
\label{fig:defense_results}
\end{figure*}

\smallskip
\noindent\textbf{Evaluating defense and rectification.}
As the defense evaluation focus is on rectifying misclassification,
our test images have a contamination level of 1---meaning all of the images are adversarial.
Comparing \method{} to adversarial training (AT), we find that utilizing robust features that semantically align with human intuition provides a significant improvement over $\gamma$-robust features learned through adversarial training.
We begin with a high-level analysis in Figure~\ref{fig:defense_results}, comparing \method{} to adversarial training (``AT'') and no defense (``None''). \method{}'s robust feature alignment performs 31.18\% better than adversarial training and 74.44\% than no defense when averaged across 8 attack vectors and all class sets (see Figure~\ref{fig:defense_results}). 

\begin{table*}[t]
\centering
\setlength{\tabcolsep}{4pt}

\begin{tabular}{llrrr|rrrrrrrrrrrr} 
\toprule
\multicolumn{2}{c}{\bfseries Setup} & 
\multicolumn{3}{c|}{\bfseries No Attk} & 
\multicolumn{3}{c}{\bfseries PGD-${L_\infty}$} & 
\multicolumn{3}{c}{\bfseries PGD-${L_\infty}$} & 
\multicolumn{3}{c}{\bfseries PGD-${L_2}$} &
\multicolumn{3}{c}{\bfseries PGD-${L_2}$} \\

\multicolumn{2}{c}{} & 
\multicolumn{3}{c|}{} & 
\multicolumn{3}{c}{$\epsilon=8$} & 
\multicolumn{3}{c}{$\epsilon=16$} & 
\multicolumn{3}{c}{$\epsilon=300$} &
\multicolumn{3}{c}{$\epsilon=600$} \\

 \cmidrule(l){1-2} \cmidrule(l){3-5} \cmidrule(l){6-8} \cmidrule(l){9-11} \cmidrule(l){12-14} \cmidrule(l){15-17}

M & CS & \textbf{None} & \s{AT} & \s{UM} & \s{None} & \s{AT} & \textbf{UM} & \s{None} & \s{AT} & \textbf{UM} & \s{None} & \s{AT} & \textbf{UM} & \s{None} & \s{AT} & \textbf{UM} \\
\midrule
\addlinespace[2ex]
\multirow{4}{*}{\rotatebox[origin=c]{90}{\textbf{ResNet}}}
& 3a & \textbf{.98} & \s{.96} & \s{.94} & \s{.31} & \s{.71} & \textbf{.85} & \s{.22} & .\s{50} & \textbf{.72} & \s{.07} & \s{.86} & \textbf{.92} & \s{.00} & \s{.67} & \textbf{.91} \\

& 3b & \textbf{.97} & \s{.94} & \s{.92} & \s{.24} & \s{.63} & \textbf{.82} & \s{.19} & \s{.47} & \textbf{.68} & \s{.01} & \s{.75} & \textbf{.89} & \s{.00} & \s{.31} & \textbf{.85} \\

& 5a & \textbf{.97} & \s{.92} & \s{.93} & \s{.17} & \s{.51} & \textbf{.82} & \s{.15} & \s{.24} & \textbf{.66} & \s{.00} & \s{.79} & \textbf{.91} & \s{.00} & \s{.57} & \textbf{.89} \\

& 5b & \textbf{.97} & \s{.92} & \s{.91} & \s{.22} & \s{.56} & \textbf{.78} & \s{.17} & \s{.34} & \textbf{.61} & \s{.04} & \s{.78} & \textbf{.88} & \s{.00} & \s{.50} & \textbf{.84} \\

\addlinespace[2ex]

\multirow{4}{*}{\rotatebox[origin=c]{90}{\textbf{DenseNet}}}
& 3a & \textbf{.97} & \s{.95} & \s{.94} & \s{.31} & \s{.70} & \textbf{.86} & \s{.24} & \s{.48} & \textbf{.74} & \s{.02} & \s{.86} & \textbf{.93} & \s{.00} & \s{.71} & \textbf{.91} \\

& 3b & \textbf{.97} & \s{.93} & \s{.92} & \s{.25} & \s{.60} & \textbf{.82} & \s{.23} & \s{.44} & \textbf{.67} & \s{.02} & \s{.79} & \textbf{.89} & \s{.00} & \s{.46} & \textbf{.85} \\

& 5a & \textbf{.97} & \s{.90} & \s{.93} & \s{.22} & \s{.51} & \textbf{.82} & \s{.18} & \s{.27} & \textbf{.66} & \s{.03} & \s{.77} & \textbf{.91} & \s{.00} & \s{.54} & \textbf{.88} \\

& 5b & \textbf{.97} & \s{.92} & \s{.91} & \s{.24} & \s{.55} & \textbf{.79} & \s{.21} & \s{.28} & \textbf{.62} & \s{.02} & \s{.81} & \textbf{.89} & \s{.00} & \s{.58} & \textbf{.85} \\

\addlinespace[2ex]
\cdashline{1-17}
\addlinespace[2ex]
\end{tabular}

\begin{tabular}{llrrr|rrrrrrrrrrrr}

\multicolumn{2}{c}{\bfseries Setup} & 
\multicolumn{3}{c|}{\bfseries No Attk} & 
\multicolumn{3}{c}{\bfseries MIA-${L_\infty}$} & 
\multicolumn{3}{c}{\bfseries MIA-${L_\infty}$} & 
\multicolumn{3}{c}{\bfseries MIA-${L_2}$} &
\multicolumn{3}{c}{\bfseries MIA-${L_2}$} \\

\multicolumn{2}{c}{} & 
\multicolumn{3}{c|}{} & 
\multicolumn{3}{c}{$\epsilon=8$} & 
\multicolumn{3}{c}{$\epsilon=16$} & 
\multicolumn{3}{c}{$\epsilon=300$} &
\multicolumn{3}{c}{$\epsilon=600$} \\

 \cmidrule(l){1-2} \cmidrule(l){3-5} \cmidrule(l){6-8} \cmidrule(l){9-11} \cmidrule(l){12-14} \cmidrule(l){15-17}

M & CS & \textbf{None} & \s{AT} & \s{UM} & \s{None} & \s{AT} & \textbf{UM} & \s{Non} & \s{AT} & \textbf{UM} & \s{None} & \s{AT} & \textbf{UM} & \s{None} & \s{AT} & \textbf{UM} \\
\midrule
\addlinespace[2ex]
\multirow{4}{*}{\rotatebox[origin=c]{90}{\textbf{ResNet}}}
& 3a & \textbf{.98} & \s{.96} & \s{.94} & \s{.00} & \s{.22} & \textbf{.82} & \s{.00} & \s{.00} & \textbf{.68} & \s{.01} & \s{.86} & \textbf{.92} & \s{.00} & \s{.68} & \textbf{.91} \\

& 3b & \textbf{.97} & \s{.94} & \s{.92} & \s{.00} & \s{.02} & \textbf{.77} & \s{.00} & \s{.00} & \textbf{.63} & \s{.00} & \s{.68} & \textbf{.89} & \s{.00} & \s{.29} & \textbf{.85} \\

& 5a & \textbf{.97} & \s{.92} & \s{.93} & \s{.00} & \s{.25} & \textbf{.79} & \s{.00} & \s{.02} & \textbf{.63} & \s{.00} & \s{.79} & \textbf{.91} & \s{.00} & \s{.58} & \textbf{.89} \\

& 5b & \textbf{.97} & \s{.92} & \s{.91} & \s{.00} & \s{.12} & \textbf{.73} & \s{.00} & \s{.01} & \textbf{.55} & \s{.01} & \s{.77} & \textbf{.87} & \s{.00} & \s{.51} & \textbf{.84} \\

\addlinespace[2ex]

\multirow{4}{*}{\rotatebox[origin=c]{90}{\textbf{DenseNet}}}
& 3a & \textbf{.97} & \s{.95} & \s{.94} & \s{.00} & \s{.37} & \textbf{.83} & \s{.00} & \s{.02} & \textbf{.69} & \s{.00} & \s{.86} & \textbf{.92} & \s{.00} & \s{.72} & \textbf{.90} \\

& 3b & \textbf{.97} & \s{.93} & \s{.92} & \s{.00} & \s{.18} & \textbf{.76} & \s{.00} & \s{.01} & \textbf{.62} & \s{.00} & \s{.78} & \textbf{.89} & \s{.00} & \s{.49} & \textbf{.85} \\

& 5a & \textbf{.97} & \s{.90} & \s{.93} & \s{.00} & \s{.27} & \textbf{.78} & \s{.00} & \s{.02} & \textbf{.58} & \s{.00} & \s{.77} & \textbf{.91} & \s{.00} & \s{.56} & \textbf{.87} \\

& 5b & \textbf{.97} & \s{.92} & \s{.91} & \s{.00} & \s{.30} & \textbf{.75} & \s{.00} & \s{.03} & \textbf{.58} & \s{.00} & \s{.80} & \textbf{.89} & \s{.00} & \s{.60} & \textbf{.84} \\

\bottomrule
\addlinespace
\end{tabular}

\caption{Accuracies in countering
4 strong attacks at 2 strength levels (PGD-$L_\infty$, PGD-$L_2$, MIA-$L_\infty$, MIA-$L_2$),
using 2 CNN architectures
as unprotected model $M$
across 4 class sets.
\method{} (``UM'') provides significantly better protection than adversarial training (``AT''), $31.18\%$ on average.
``None'' means no defense.
}
\label{table:unmask_results}
\end{table*}

In Table~\ref{table:unmask_results}, we analyze the information contained in Figure~\ref{fig:defense_results} in detail.
One key observation is that \textit{feature overlap} is a dominant factor in determining the accuracy of the \method{} defense, as opposed to the number of classes. When examining the ResNet50 model on MIA-$L_\infty$ ($\epsilon$=8), class set CS3b (3 classes; feature overlap 50\%), \method{} is able to determine the underlying class 77\% of the time. At class set CS5a (5 classes; feature overlap 23.53\%) an accuracy of 79\% is obtained, highlighting the important role that feature overlap plays in \method's defense ability. Similar trends can be observed across many of the attacks.
In addition, Table~\ref{table:unmask_results} highlights that \method{} is agnostic to the deep learning model that is being protected (ResNet50 vs DenseNet121), as measured by the ability of \method{} to infer an adversarial images' actual class.

It is interesting to observe that MIA-$L_\infty$ is more effective at breaking the \method{} defense. 
We believe this could be due to the single-step attacks' better transferability, which has been reported in prior work~\cite{advtraining1}. 
We also note the fact that \method's accuracy could be higher than the un-attacked model $M$ if model $K$ learns a better representation of the data through the feature masks as opposed to model $M$, which trains on the images directly.

\begin{figure*}[t]
\centering
\includegraphics[width=0.8\textwidth]{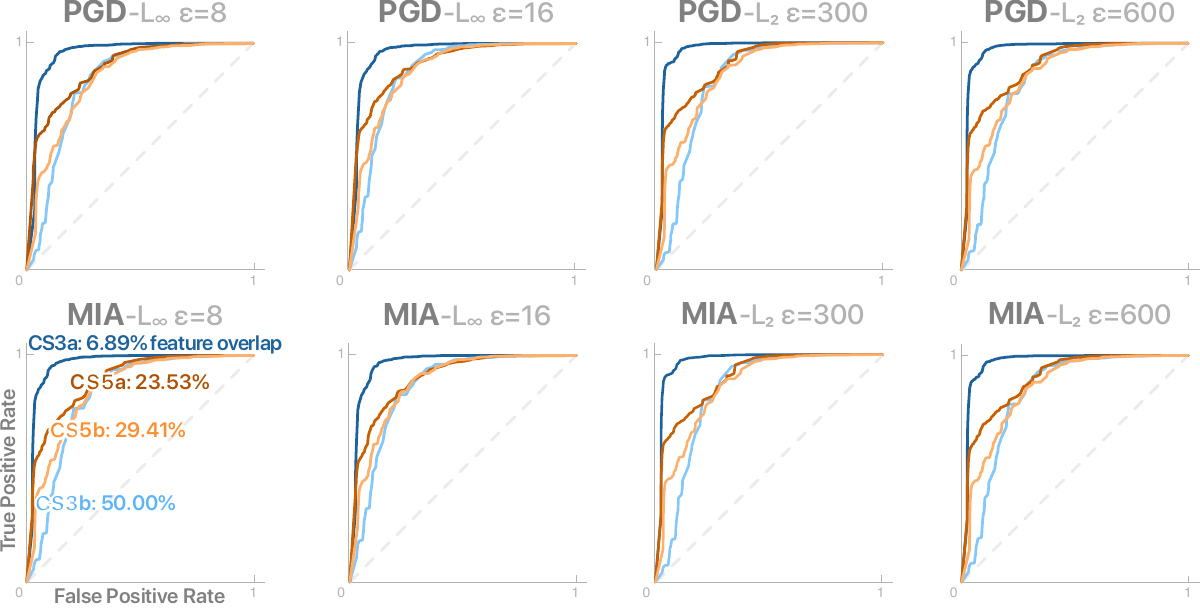}
\caption{\method's effectiveness in detecting 4 strong attacks at two strength levels.
\method{}'s protection may not be affected strictly based on the number of classes.
Rather, an important factor is the \textit{feature overlap}  among classes.
\method{} provides better detection when there are 5 classes (dark orange; 23.53\% overlap) than when there are 3 (light blue; 50\% overlap).
Keeping the number of classes constant and varying their feature overlap also supports our observation about the role of feature overlap (e.g., CS3a at 6.89\% vs. CS3b at 50\%).
Dotted line indicates random guessing.
}
\label{roc_curves}
\end{figure*}

\smallskip
\noindent\textbf{Evaluating attack detection.}
To evaluate \method{}'s effectiveness in detecting adversarial images,
we set the contamination level to 0.5---meaning half of the images are benign and the other half are adversarial.
Figure \ref{roc_curves} summarizes \method{}'s detection effectiveness, using \textit{receiver operating characteristics} (ROC) curves constructed by varying the adversarial-benign threshold $t$.
The curves show \method{}'s performance across operating points as measured by the tradeoff between \textit{true positive} (TP) and \textit{false positive} (FP) rates. 
Table~\ref{table:unmask_detection} shows the number of images used to test the detection ability of \method{}. Only images that are successfully attacked are used for evaluation (combined with benign counterparts), thus the variations in numbers.

An interesting characteristic of \method{}'s protection is that its effectiveness may not be affected strictly based on the number of classes in the dataset as in conventional classification tasks.
Rather, an important factor is how much \textit{feature overlap} there is among the classes.
The ROC curves in Figure~\ref{roc_curves} illustrate this phenomenon, 
where \method{} provides better detection when there are 5 classes (Figure~\ref{roc_curves}, dark orange) than when there are 3 classes (light blue).
As shown in Table~\ref{table:unmask_dataset}, 
the 5-class setup (CS5a---dark orange) has a feature overlap of 23.53\% across the the 5 classes' 34 unique features, while the 3-class setup (CS3b---light blue) has 50\% overlap.
Keeping the number of classes constant and varying their feature overlap also supports this observation about the role of feature overlap (e.g., CS3a vs. CS3b in Figure~\ref{roc_curves}).

For a given feature overlap level, 
\method{} performs similarly across attack methods.
When examining feature overlap 6.89\% (CS3a) on DenseNet121, 
\method{} attains AUC scores of 0.95, 0.958, 0.968, 0.967, 0.962, 0.961, 0.969 and 0.967 on attacks PGD-$L_\infty$ ($\epsilon$=8/16), PGD-$L_2$ ($\epsilon$=300/600), MIA-$L_\infty$ ($\epsilon$=8/16) and MIA-$L_2$ ($\epsilon$=300/600), respectively.
This result is significant because it highlights the ability of \method{} to operate against multiple strong attack strategies to achieve high detection success rate. 
As a representative ROC operating point for the attack vectors, we use MIA-$L_2$ ($\epsilon$=300) on feature overlap 6.89\%. 
In this scenario, \method{} is able to detect up to 96.75\% of attacks with a false positive rate of 9.66\%.  
We believe that performing well in a low feature overlap environment is all that is required. 
This is because in many circumstances it is not important to distinguish the exact class (e.g., dog or cat) of the image, but whether the image is being completely misclassified (e.g., car vs. person). 
Therefore, in practice, classes can be selected such that feature overlap is minimized.

\section{Conclusion}
In this paper, we have introduced a new method for semantically aligning robust features with human intuition, and showed how it protects deep learning models against adversarial attacks through the \method{} detection and defense framework.
Through extensive evaluation, we analyze the merits of \method's ability to \textit{detect} attacks---finding up to 96.75\% of attacks with a false positive rate of 9.66\%; and \textit{defend} deep learning models---correctly classifying up to 93\% of adversarial images in the gray-box scenario.
\method{} provides significantly better protection than adversarial training across 8 attack vectors, averaging 31.18\% higher accuracy.
Our proposed method is fast and architecture-agnostic.
We expect our approach to be one of multiple techniques used in concert to provide comprehensive protection.
Fortunately, our proposed technique can be readily integrated with many existing techniques, as it operates in parallel to the deep learning model that it aims to protect.

\section{Acknowledgements}
This work was in part supported by NSF grant IIS-1563816, CNS-1704701, GRFP (DGE-1650044), a Raytheon research fellowship and a DARPA grant. 
Use, duplication, or disclosure is subject to the restrictions as stated in Agreement number HR00112030001 between the Government and the Performer.

\bibliography{main.bib}
\bibliographystyle{IEEEtran}

\end{document}